# Does YOLO26 Truly Offer Advantages Over Its Predecessors for Edge Deployment? A Benchmark Study in Aquaculture

Rakesh Ranjan[1,*], Gajanan S. Kothawade[1], Kata Sharrer[1], Scott Tsukuda[1], Christopher Good[1]

[1]The Conservation Fund Freshwater Institute, Shepherdstown, WV, 25443

[*]Corresponding author, Email: rranjan@conservationfund.org, Phone: +1 (304) 870-2203

**Abstract**

The You Only Look Once (YOLO) has been widely adopted in aquaculture monitoring and management due to its real-time performance and deployment flexibility. The recently introduced YOLO26 architecture incorporates Non-Maximum Suppression (NMS)-free end-to-end inference and is optimized for deployment on resource-constrained CPU-based devices, making it particularly relevant for edge deployment in commercial aquaculture applications. Nevertheless, its performance, operational efficiency, and deployment suitability compared with previous YOLO generations remain largely unvalidated in aquaculture-specific scenarios. This study benchmarks YOLO26 against three Ultralytics predecessors (YOLOv5u, YOLOv8, and YOLO11) across nano, small, and medium model scales for the detection of fish mortality, a critical indicator of fish population health and welfare, in recirculating aquaculture systems (RAS). Twelve model variants were evaluated for detection accuracy, training efficiency across seven dataset sizes, and inference performance on both high-performance NVIDIA A100 GPUs and the resource-constrained, CPU-only Raspberry Pi 5 edge device. All models achieved comparable performance on the full dataset, with mAP50 varying by only 1.04 percentage points, indicating minimal influence of architectural generation on final mortality detection accuracy when sufficient training data are available. However, notable differences emerged in data efficiency and deployment performance. YOLOv8 demonstrated the strongest training efficiency, achieving 90% mAP50 with only 400 training images, whereas YOLO26 nano and small variants required 1,000 images to reach comparable accuracy. In contrast, YOLO26 exhibited advantages during edge deployment, with YOLO26n achieving the highest inference speed on the Raspberry Pi 5 at 7.51 FPS, while YOLOv5mu outperformed all contemporary medium-scale architectures on CPU-based hardware. These results demonstrate that architectural novelty alone is an insufficient criterion for model selection. The findings support a deployment-oriented framework in which training data availability, target hardware, and inference requirements collectively inform model selection for aquaculture applications.

## 1. Introduction

With the growing global population and rising per capita seafood consumption, demand for aquaculture production continues to increase [1]. Intensive aquaculture systems, such as net pens and recirculating aquaculture systems (RAS), have shown promise to meet this growing demand by supplying affordable, high-quality sea food [2,3]. However, intensive systems also introduce significant challenges related to sustainability, fish health and welfare, and environmental impacts [4,5]. For RAS, the high capital and operational costs of these systems place strong pressure on profit margins and demand a high resource efficiency to ensure economic viability [6]. Recent advances in precision aquaculture technology, particularly Computer Vision and Artificial Intelligence (AI)-aided automated monitoring and decision-support systems, offer a promising tool for mitigating operational challenges and maximizing yield in intensive production environments [7,8]. These methods enable automated, non-invasive, and continuous monitoring of key biological and operational parameters in aquaculture systems that were previously accessible only through costly, labor-intensive manual observations.

Several studies have reported the adoption of convolutional neural networks (CNNs) for computer vision applications in aquaculture [9,10]. CNNs are machine learning models specifically designed for processing image and video data and are widely used for object detection and instance segmentation tasks. CNN-based methods have been applied to a wide range of aquaculture applications, including feed optimization [9,11], animal behavior monitoring [12,13], biomass estimation [14], disease detection [15], mortality monitoring [16,17], and system inspection [18]. While domain-specific adaptation of CNNs is being increasingly researched, their deployment in intensive aquaculture faces unique challenges, including high stocking densities, elevated water turbidity, optical distortions, and the rapid, unpredictable motion of aquatic animals. Unlike stationary targets in crop-based applications, key objects in aquaculture, such as small uneaten feed pellets randomly floating in a highly occluded complex background, complicate object detection and tracking. Furthermore, remote aquaculture facilities often face constraints on high-bandwidth data transmission, which limits the feasibility of cloud-based CNN deployment. Therefore, deploying CNN models on resource-constrained edge devices is critical in aquaculture to enable real-time, on-site monitoring and decision support without relying on continuous cloud connectivity or high-bandwidth communication infrastructure.

Inference speed is a key criterion in selecting CNN models for real-time aquaculture monitoring applications, such as tracking morphological features and welfare-related traits in freely swimming fish. While early two-stage CNN object detectors (e.g., R-CNN, Faster R-CNN, and Mask R-CNN) achieve high detection accuracy, their sequential pipelines introduce substantial computational overhead, resulting in increased latency and reduced inference speed [19,20]. These models first generate candidate Regions of Interest (ROIs) using a Region Proposal Network (RPN), followed by classification, bounding-box regression, and confidence re-scoring to produce final detections. A state-of-the-art (SOTA) two-stage Faster R-CNN achieves only 7–18 FPS on a Titan X GPU [21]. On embedded devices such as the Jetson TX2, reported inference speeds fall below 1 FPS, making these models unsuitable for real- or near-real-time detection in edge applications [22]. To address these challenges, single-stage CNN object detectors (e.g., EfficientDet, RetinaNet, SSD, YOLO) were introduced. These models eliminate the region proposal generation step in detection and perform object classification and bounding box regression simultaneously in a single forward pass [23]. Among these models, You Only Look Once (YOLO) has gained popularity in computer vision research for its optimal balance of detection accuracy and inference speed. The first YOLO model (YOLOv1), introduced by Redmon et al. [21] achieved real-time object detection with inference speeds of 45 FPS for the base model and up to 155 FPS for Fast YOLO on a Titan X GPU. This represented the first unified object detection framework to achieve real-time performance while maintaining decent accuracy. While YOLOv1 achieved superior speed, its accuracy was not on par with that of the two-stage Faster R-CNN model. The subsequent YOLO versions (e.g., YOLOv3, YOLOv5, YOLOv8), however, further optimized the architecture to achieve detection performance comparable to SOTA two-stage CNN models [24–26]. Unlike other detection frameworks that have evolved primarily within academic research groups, YOLO has undergone rapid, community-driven iteration, from YOLOv1 through YOLO26, with each version introducing architectural innovations that have been quickly adopted, modified, and validated across applied domains, including aquaculture.

Mortality is a key metric for assessing the health and welfare of aquaculture populations. Unusual mortality patterns may indicate disease outbreaks, environmental stress, or system failures, among other causes. Real- or near-real-time mortality tracking can provide farm managers with actionable insights to inform management decisions and mitigate root causes before mass mortality events occur. Several research studies have refined the original YOLO architecture and introduced custom frameworks such as D-YOLOv4/E-YOLOv4 [27], Improved YOLOv8 [28], and Deadfish-YOLO [17] to effectively address domain-specific challenges in mortality detection. The incorporation of attention mechanisms, lightweight backbone architectures, redesigned loss functions, and transformer modules into optimized detection frameworks has improved the model's detection performance. Nevertheless, deployment-oriented performance analysis of YOLO models and the practical challenges in resource-constrained edge deployments remain largely underexplored [29]. The majority of reported studies benchmark detection performance exclusively on GPU-equipped servers or laboratory workstations, which do not reflect the constraints of real-world aquaculture deployments. Although GPU-enabled embedded platforms (e.g., NVIDIA Jetson) offer on-site inference capabilities, they remain cost-prohibitive and consume relatively high power, limiting their scalability for per-tank or per-pen deployment in commercial operations. Low-power embedded systems, such as Raspberry Pi and other CPU-based single-board computers, are cost-effective and well-suited for real-world deployment and increasingly adopted in small- to medium-scale aquaculture settings for sensing and automation tasks [30,31].

While most CNN-based detection models are optimized for GPU acceleration, recent advances in the YOLO family, particularly the latest YOLO26, have introduced streamlined architectures that enable faster, more resource-efficient inference on CPU-constrained edge devices [32]. YOLO26 has been reported to achieve improved small-object detection accuracy and up to 43% faster CPU inference compared to its predecessor, YOLO11 [33]. However, these performance gains have been demonstrated primarily on general-purpose benchmarks such as the Common Objects in Context (COCO) dataset, and their applicability to domain-specific aquaculture tasks has not been validated in the peer-reviewed literature. There is a critical gap in evaluating the deployment feasibility, performance trade-offs, and practical limitations of modern lightweight detection architectures under aquaculture-specific conditions on resource-constrained edge hardware. To address this gap, this study conducts a deployment-oriented benchmark that systematically evaluates YOLO26 against prior Ultralytics YOLO models in terms of detection accuracy, data efficiency, inference speed, and real-world edge-deployment feasibility for fish mortality monitoring in RAS. Given the significance of mortality monitoring in aquaculture, this study leverages mortality data collected from a semi-commercial scale RAS system to ensure real-world validation. The specific objectives of this study are to:

1. Quantify and compare the detection accuracy and model efficiency of YOLO26 for fish mortality detection against other Ultralytics YOLO architectures (YOLOv5, YOLOv8, and YOLOv11) across various relevant model scales.
2. Assess training efficiency by determining the minimum data requirements for each architecture using learning curve analysis.

3. Benchmark inference performance on both high-performance GPU and resource-constrained CPU-based edge hardware to assess the practical deployment feasibility of each architecture for on-farm RAS mortality monitoring.

## 2. Materials and Methods

### 2.1 YOLO model architecture

YOLO has established itself as the standard for deployment-oriented computer vision applications across diverse domains, including agriculture and aquaculture. Ultralytics is a leading developer and maintainer of modern YOLO architectures and has played a key role in standardizing YOLO as a unified, production-ready framework to enhance accessibility, reproducibility, and adoption for real-time object detection. To ensure fair benchmarking of accuracy, efficiency, and edge-deployment performance, this study exclusively evaluates the Ultralytics YOLO (YOLOv5u [25], YOLOv8 [26], YOLO11 [33], and YOLO26 [32]) architectures, as these models share a consistent framework, training, and implementation pipeline. Non-Ultralytics and legacy YOLO variants were excluded to avoid inconsistencies. The details of key architectural innovations and contributions for each Ultralytics YOLO have been summarized in Table 1. While YOLOv5u, YOLOv8, and YOLO11 have been extensively documented in the literature and their architectural details are well established [34–36], YOLO26 represents the most recent addition to the Ultralytics family, with limited peer-reviewed coverage and no reported application in the aquaculture domain to date.

YOLO26 [32,35] is the latest-generation Ultralytics architecture, officially released in January 2026. The model is explicitly engineered for edge and low-power devices, featuring a streamlined design that reduces architectural complexity while incorporating task-specific optimizations to enable faster, lighter, and more accessible deployment. YOLO26 is a native end-to-end model that eliminates the Non-Maximum Suppression (NMS) post-processing step, resulting in lighter and faster inference (Figure 1). This is particularly beneficial for CPU-based edge deployment, where the sequential sorting and filtering operations inherent to NMS introduce greater computational overhead than on GPU hardware. The Distribution Focal Loss (DFL) module is also removed to simplify the bounding box regression pipeline, enabling seamless export to ONNX, CoreML, TensorRT, and TFLite formats for efficient deployment on resource-constrained edge devices. Training stability and detection accuracy are further enhanced by the MuSGD optimizer, which promotes faster convergence, along with ProgLoss (progressive loss balancing) and STAL (Small-Target-Aware Label Assignment) to improve the detection of small and spatially sparse objects. These optimizations collectively improve small-object detection accuracy, simplify deployment across a broad range of hardware, and deliver CPU inference up to 43% faster than YOLO11 on the COCO benchmark [32]. YOLO26 supports real-time performance across multiple vision tasks, including object detection, instance segmentation, classification, pose estimation, and oriented object detection.

YOLO26 was selected as the primary model for this benchmark study because its architecture is theoretically well-suited for deployment on resource-constrained on-farm hardware and for addressing real-world mortality detection challenges, including detecting overlapping targets during high-mortality events (Figure 1).

**Table 1:** Summary of Ultralytics YOLO architectures evaluated in this benchmark study.

| Model (Year) | Key architectural innovation and contribution | Tasks | Framework | Citation |
|---|---|---|---|---|
| YOLOv5u* (2020) | Ultralytics re-implementation of YOLOv5 with an anchor-free detection head and decoupled classification/regression branches. Employs CSP (Cross-Stage Partial) backbone with SiLU activation and PANet neck for multi-scale feature aggregation. The 'u' suffix denotes the updated anchor-free variant used in this study. | Object Detection, Instance Segmentation | PyTorch (Ultralytics) | [25] |
| YOLOv8 (2023) | Ultralytics next-generation model featuring a new C2f backbone and a fully anchor-free, decoupled detection head. Introduced improved training strategies, including mosaic augmentation, closure, and dynamic task-aligned assignment. Supports multi-task learning natively across detection, segmentation, pose estimation, and classification. | Object Detection, Instance Segmentation, Pose Estimation, Classification | PyTorch (Ultralytics) | [26] |
| YOLO11 (2024) | Introduced C3k2 CSP bottleneck blocks throughout backbone and neck for enhanced efficiency with fewer parameters than YOLOv8. Incorporated C2PSA (CSP with Spatial Attention) module to improve feature focus on salient regions. Achieves 22% fewer parameters than YOLOv8m while maintaining higher COCO mAP. Extended YOLO to oriented detection. | Object Detection, Instance Segmentation, Pose Estimation, Oriented Detection, Classification | PyTorch (Ultralytics) | [33,36] |
| YOLO26 (2026) | Edge-optimized end-to-end model eliminating Non-Maximum Suppression (NMS) via native one-to-one label assignment, removing post-processing latency. Removed Distribution Focal Loss (DFL) for simpler, hardware-friendly bounding box regression. Introduced MuSGD optimizer (SGD+Muon hybrid) for stable convergence, ProgLoss for progressive loss balancing, and Small-Target-Aware Label Assignment (STAL) for improved small-object detection accuracy. Delivers up to 43% faster CPU inference than YOLO11 on edge devices. | Object Detection, Instance Segmentation, Pose Estimation, Oriented Detection, Classification | PyTorch (Ultralytics) | [32,35] |

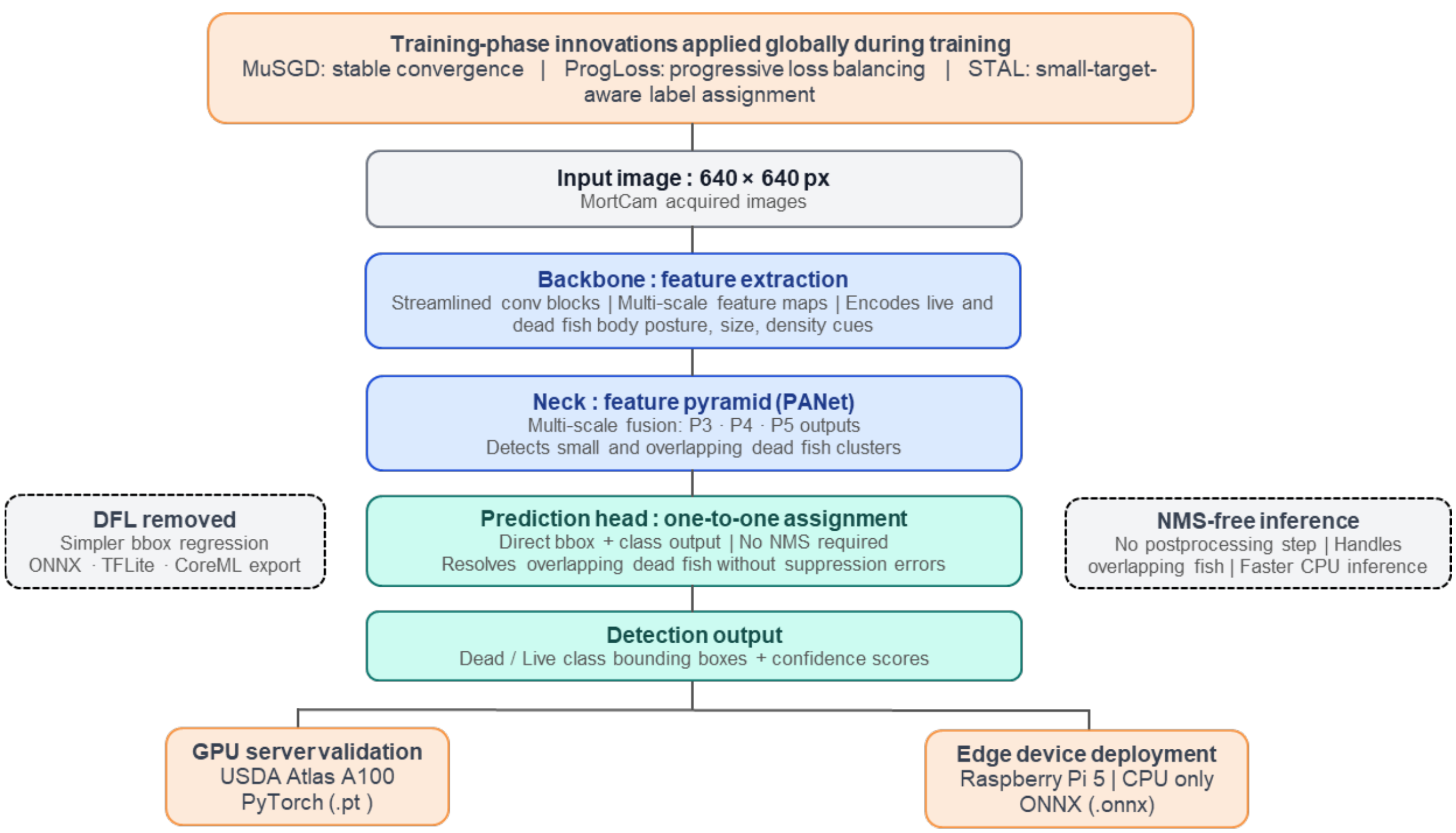

Figure 1. YOLO26 architecture, illustrating the end-to-end inference pipeline and key architectural innovations. The mortality model was trained and benchmarked on an NVIDIA A100 GPU (USDA SCINet Atlas) and deployed on a Raspberry Pi 5 to evaluate real-world inference performance on a resource-constrained edge device [NMS: Non-Maximum Suppression; DFL: Distribution Focal Loss].

## 2.2 Dataset

### *2.2.1 Image acquisition*

The mortality data were collected in a semi-commercial-scale RAS grow-out tank (volume = 150 $m^3$) stocked with 5,000 Atlantic salmon (*Salmo salar;* stocking density = 54.3 kg $m^{-3}$; mean fish weight = 1.63 kg) at The Conservation Fund Freshwater Institute (Shepherdstown, WV, USA). A custom-built mortality imaging system (MortCam) [16] equipped with an RGB sensor (RPi HQ Camera, Raspberry Pi Foundation, Cambridge, UK; 12.3 MP resolution), a wide-angle lens (M23272M14, Arducam, Nanjing, China; 2.72 mm focal length, F2.5 aperture, 140° horizontal field of view), and an integrated single-board computer (Raspberry Pi 4B; 1.5 GHz quad-core Cortex-A72 processor) was used for data collection (Figure 2). MortCam was deployed vertically downward, suspended from a steel cable at the center of the tank, positioned 3 m below the water surface and 0.6 m above the bottom drain plate. The camera was programmed to capture still images every 15 minutes. A pair of waterproof LED lights (LUMEN-LIGHT-R4-RP, Blue Robotics Inc.; pressure rating: 500 m; maximum brightness: 1,500 lumens) was mounted on the MortCam housing to enhance illumination at the tank bottom. For the first 45 days, images were acquired under ‘ambient light’, with only the ceiling LED lights (model DLE-18-ST-W-5000–00, Digital Lumens, Boston, Massachusetts, USA) installed 7.9 m above the drain plate. Thereafter, for the next 45 days, underwater LED lights were switched on to capture images in ‘supplemental light’ conditions. During the data-collection window, mortality in the grow-out tank was marginal, and the acquired dataset couldn’t capture high-mortality scenarios. To deal with this, euthanized culls from the same population were added to the tank on the last day to simulate a high-mortality event, and images were captured under both lighting conditions. Further details on MortCam development, installation, and data acquisition protocol can be found in a study by Ranjan et al. [16].

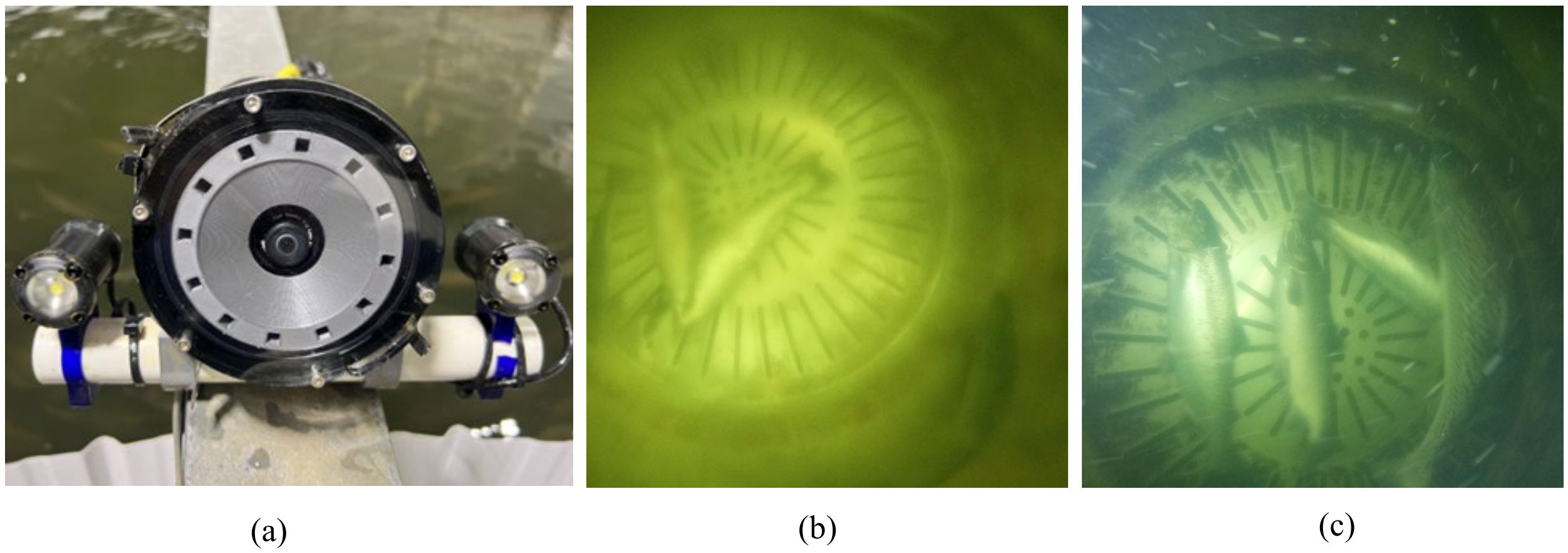

(a) (b) (c)

Figure 2. The (a) MortCam was installed above the bottom drain plate of a semi-commercial-scale grow-out RAS tank to image live and dead fish within the camera's field of view under (b) ambient and (c) supplemental lighting conditions.

*2.2.2 Data preprocessing*

A total of 2,000 images were manually sorted and selected for model training, representing diverse mortality levels [zero, low (<3 dead fish), and high (≥3 dead fish)], illumination conditions (ambient and supplemental), and varying water quality scenarios captured over three months of camera deployment. These images were uploaded to Roboflow software (Des Moines, Iowa, USA) for data annotation and preprocessing. The images were manually annotated into two object classes ('live': actively swimming fish; 'dead': fish exhibiting belly-up floating behavior indicative of mortality) and randomly split in a 70:20:10 ratio for model training, validation, and testing. Each of the datasets was preprocessed in two steps: 1) auto-orientation by removing EXIF data, and 2) image resizing (416 × 416 pixels). Thereafter, the preprocessed training set was augmented [dataset doubled (2×) using adjustment to brightness ( ± 25 %), rotation ( ± 90º), and blur ( ± 2 pixels)] using Roboflow's built-in augmentation pipeline, yielding a final training set of 2,800 images. The training dataset had 2,204 instances of the 'dead' class and 3,209 instances of the 'live' class (Table 2). The validation and test sets were retained in their original form to provide unbiased performance estimates. All experiments used these fixed splits to ensure comparability across model evaluations.

Table 2. Summary of the mortality dataset used in this benchmark study

| **Split (%)** | **Original Images** | **After Augmentation** | **Dead Instances** | **Live Instances** |
|---|---|---|---|---|
| Training (70%) | 1,400 | 2,800 (2×) | 2,204 | 3,209 |
| Validation (20%) | 400 | 400 (none) | 554 | 734 |
| Test (10%) | 200 | 200 (none) | 311 | 310 |

| Total | 2,000 | 3,400 | 3,069 | 4,253 |
|---|---|---|---|---|

## 2.3 Model training

Twelve object detection models from four Ultralytics YOLO generations (YOLOv5, YOLOv8, YOLOv11, and YOLO26) across three size tiers [nano (n), small (s), and medium (m)] were selected for this benchmark study. Larger model variants [large (l), and extra-large (x)] were excluded due to higher computational demands, larger memory footprint, and slower inference speeds, which limit their suitability for edge deployment scenarios. Each model was initialized with weights pretrained on the MS COCO dataset prior to fine-tuning.

### *2.3.1 Computing hardware*

All model training was conducted on the USDA Agricultural Research Service SCINet Atlas high-performance computing cluster (Mississippi State University, Starkville, MS, USA) with one NVIDIA A100 GPU (80 GB VRAM) and four CPU tasks per allocation. Interactive training sessions were accessed through Atlas Open OnDemand (OOD) using JupyterLab in a custom Conda environment. The software stack included Python (v3.10.20), PyTorch (v2.10.0) with CUDA 12.8 support, and Ultralytics (v8.4.21) for YOLO model training and validation.

### *2.3.2 Hyperparameter tuning*

All models were trained under standardized hyperparameter settings to ensure fair cross-model comparison. Pretrained MS COCO weights were used for initialization across all architectures. Training was conducted for up to 100 epochs with an input image size of 640 pixels, a batch size of 32, 4 CPU data-loading workers, and an early-stopping patience of 50 epochs. Optimization parameters, including optimizer selection and learning rate scheduling, were set to Ultralytics' default automatic settings. The default Ultralytics augmentation pipeline was applied uniformly, incorporating mosaic augmentation, mixup, random flipping, HSV color jitter, and scale/translation transformations.

### *2.3.3 Learning curve analysis*

Learning curve analysis was performed to assess data efficiency, performance saturation, and the minimum dataset size needed for reliable accuracy on the edge devices. The analysis was performed across 12 selected models (4 model generations × 3 size tiers) using 7 training dataset sizes (100, 200, 400, 700, 1,000, 1,400, and 2,800 images), resulting in 84 total training runs. Stratified random subsets were sampled from the entire 2,800-image training set (hereafter termed as ‘full dataset’) using a fixed seed of 42 to generate datasets of different sizes. Since each original image had a corresponding augmented version, image pairs were sampled together to preserve augmentation balance across all subset sizes. The validation and test sets were kept constant across all experiments to ensure a fair performance comparison.

## 2.4 Model performance on GPU

The detection performance of all models was evaluated using mean average precision at an IoU threshold of 0.50 (mAP50), mean average precision across IoU thresholds from 0.50 to 0.95 in 0.05 increments (mAP50–95), precision, and recall [34]. Additionally, inference speed was assessed on the training GPU using two complementary metrics. First, raw GPU inference time was measured as the neural network forward-pass time (*Infer*, ms) per image using the Ultralytics validation pipeline on the test set. This metric reflects the true computational cost of the model by excluding system-level overhead and input/output (I/O) processes. Second, end-to-end average latency (*Avg*, ms) was evaluated to capture the full per-image inference pipeline, including image loading, preprocessing (resizing and normalization), forward-pass computation, and post-processing (e.g., NMS or equivalent). This metric was computed in Python across all 200 test images after three warm-up iterations to mitigate initialization effects and ensure stable timing estimates. The corresponding GPU throughput was then derived as the frame rate (GPU FPS) from the measured end-to-end latency. The postprocessing time (*Post*, ms) was reported separately to distinguish NMS overhead between NMS-based (YOLOv5u, YOLOv8, YOLO11) and NMS-free (YOLO26) architectures. Additionally, training time (hours, h), number of parameters (millions, M), PyTorch (PT) model size (MB), and ONNX export size (MB) were recorded as supplementary efficiency indicators. Model size can be a limiting factor for deployment on edge devices with limited memory and storage. Therefore, model efficiency was quantified as the mAP50 per megabyte of PyTorch (PT) model size (×100) to evaluate the accuracy-to-storage trade-off.

## 2.5 Edge deployment evaluation

Edge inference benchmarking was performed on a Raspberry Pi 5 (Broadcom BCM2712), featuring a quad-core ARM Cortex-A76 CPU at 2.4 GHz and 8 GB LPDDR4X RAM, running Debian GNU/Linux 12 (Bookworm). The device was configured with Python (v3.13.5), Ultralytics (v8.4.23), ONNX Runtime (CPU), PyTorch (CPU build, version 2.10.0+cpu), and psutil (v5.9.4) for deployment evaluation. No GPU-accelerated co-processor was used with this device. An active cooling system comprising an integrated fan, aluminum heatsink, and thermal pads was installed to mitigate thermal throttling during sustained inference workloads. The trained weights of 12 full dataset models were exported from PyTorch (.pt) to the Open Neural Network Exchange (ONNX) format on the Atlas cluster. The ONNX models, along with the 200 test images, were downloaded to a host computer and subsequently transferred to the edge device. Exported ONNX model sizes ranged from 9.8 MB (YOLO26n) to 103.6 MB (YOLOv8m). During inference benchmarking, device temperature and thermal throttling status were verified prior to each model evaluation. Three warm-up inference runs were performed on the first test image to initialize the ONNX Runtime execution environment. Evaluated inference metrics include mean inference time (*Avg*, ms), minimum (*Min*, ms), maximum (*Max*, ms), and 90th percentile (*P90*, ms) latency, along with frames per second (FPS), and relative percentage FPS compared to the baseline YOLOv8 model.

The mortality dataset (*RAS-MortDB*) and trained model weights in PyTorch (.pt) and ONNX (.onnx) formats for all twelve evaluated architectures will be publicly available at GitHub to support reproducibility and advance open-access aquaculture computer vision research.

## 3. Results and Discussion

### 3.1 Mortality detection

#### *3.1.1 Overall detection accuracy*

The detection performance of 12 YOLO models trained on the full dataset has been presented in Table 3. All models achieved competitive accuracy, with mAP50 values ranging from 93.77% (YOLOv8m) to 94.81% (YOLO11m), representing a narrow spread of only 1.04 percentage points across all architectures and size tiers. Similarly, the mAP50-95 metric, which applies stricter IoU thresholds, ranges from 68.48% (YOLOv5nu) to 72.12% (YOLO26s), indicating generally robust bounding-box localization across all models. The marginal differences in detection performance across the tested models and tiers indicate that, for the binary mortality detection task (dead vs. live), where dead fish exhibit strong visual indicators such as a belly-up posture and cessation of movement, architecture selection has minimal influence on overall detection performance. Even the smallest nano-tier models fully capture the discriminative features with adequate training data (Figure 3). The practical relevance of mAP50-95 in the context of RAS mortality monitoring is noteworthy. While loose detection at IoU of 50% may suffice for mortality count estimation, tighter bounding-box precision is important for distinguishing individual carcasses in high mortality events, where overlapping boxes can negatively affect detection count accuracy. Precision and recall metrics followed similar patterns, with minimal variation across architectures and model sizes. Overall, these findings indicate that newer-generation YOLO architectures, including the recently introduced YOLO26, do not offer a substantial advantage in detection accuracy for this specific application. Despite YOLO26's advanced loss optimization strategies, including ProgLoss and STAL, which are intended to enhance spatial precision for small and densely clustered objects, these improvements did not translate into measurable performance gains for mortality detection in larger fish. Notably, this benchmark dataset primarily comprised larger fish (1.5–2.5 kg), and juvenile mortality scenarios were not represented. Consequently, further investigation is needed to determine whether newer YOLO architectures may provide meaningful advantages for juvenile mortality detection, where smaller body size and potentially more challenging visual characteristics may alter model performance differentials.

Table 3. Detection performance of YOLO models trained on the full dataset (2,800 images).

| Model | Tier | mAP50 (%) | mAP50-95 (%) | Precision (%) | Recall (%) |
|---|---|---|---|---|---|
| YOLO26n | Nano | 94.32 | **71.27** | **90.79** | 89.29 |
| YOLO11n | | 94.21 | 70.93 | 88.75 | **89.72** |
| YOLOv8n | | **94.51** | 69.4 | 90.52 | 88.61 |
| YOLOv5nu | | 94.1 | 68.48 | 90.62 | 88.25 |
| YOLO26s | Small | **94.47** | **72.12** | 89.21 | **90.41** |
| YOLO11s | | 94.46 | 71.96 | 89.54 | 89.96 |
| YOLOv8s | | 94.46 | 69.7 | 89.75 | 89.08 |
| YOLOv5su | | 94.03 | 70.34 | **91.22** | 87.74 |
| YOLO26m | Medium | 94.40 | **71.81** | **90.46** | 88.97 |
| YOLO11m | | **94.81** | 71.63 | 90.07 | 88.07 |
| YOLOv8m | | 93.77 | 71.19 | 89.16 | 87.65 |
| YOLOv5mu | | 94.65 | 70.97 | 89.64 | **90.67** |

*3.1.2 Tier-level performance analysis*

Within the nano tier, YOLOv8n achieved the highest mAP50 at 94.51%, marginally outperforming YOLO26n (94.32%), YOLO11n (94.21%), and YOLOv5nu (94.10%) [Table 3]. YOLO26s, YOLO11s, and YOLOv8s had almost identical mAP50 scores (94.46–94.47%), followed by YOLOv5su (94.03%) in the small tier. For the medium tier, YOLO11m achieved the highest accuracy at 94.81%, followed by YOLOv5mu (94.65%), YOLO26m (94.40%), and YOLOv8m (93.77%). The marginal accuracy advantage of one model over another for all three size tiers falls well within normal run-to-run training variance and should not be interpreted as a practically meaningful performance difference for on-farm deployment. These findings further indicate that the model architecture and tier selection do not affect the object detection accuracy in this specific application.

*3.1.3 Per-class detection analysis*

Across all 12 evaluated models, detection performance for the Dead class (mAP50: 95.1–96.0%) consistently exceeded the Live class (mAP50: 91.9–93.7%). This trend is particularly noteworthy given that Live instances were more abundant in the training dataset, indicating that class frequency alone did not determine detection success. The comparatively inferior performance in Live fish detection is primarily attributable to greater visual variability and image-quality challenges associated with live fish behavior. Live fish exhibited rapid, continuous, and unpredictable movement around the bottom drain plates, frequently resulting in motion blur observed in the training images that reduced feature clarity. Moreover, live fish often swam close to the camera, causing partial body captures and incomplete object representation within the frame. These factors introduced substantial within-class variability in orientation, body position, depth, and motion state, thereby increasing the complexity of precise object localization and bounding-box assignment. In contrast, Dead fish displayed slower, more consistent movement driven primarily by water currents and were typically concentrated near the bottom drain plate region, which remained consistently within the camera's field of view. As a result, images of dead fish were generally clearer, with more complete body visibility and reduced positional variability. The higher visual consistency and improved image quality of Dead class samples likely contributed to the superior detection accuracy observed across all models. This per-class performance pattern is consistent with previous findings reported by Ranjan et al. [16], who similarly observed stronger detection confidence for Dead class instances in a YOLOv7-based RAS mortality monitoring system. From an operational perspective, the uniformly high Dead class detection accuracy across all architectures suggests that each evaluated model can provide reliable mortality alert functionality. Therefore, for practical deployment, model selection may be more appropriately guided by computational efficiency and inference speed requirements rather than marginal differences in class-specific detection performance.

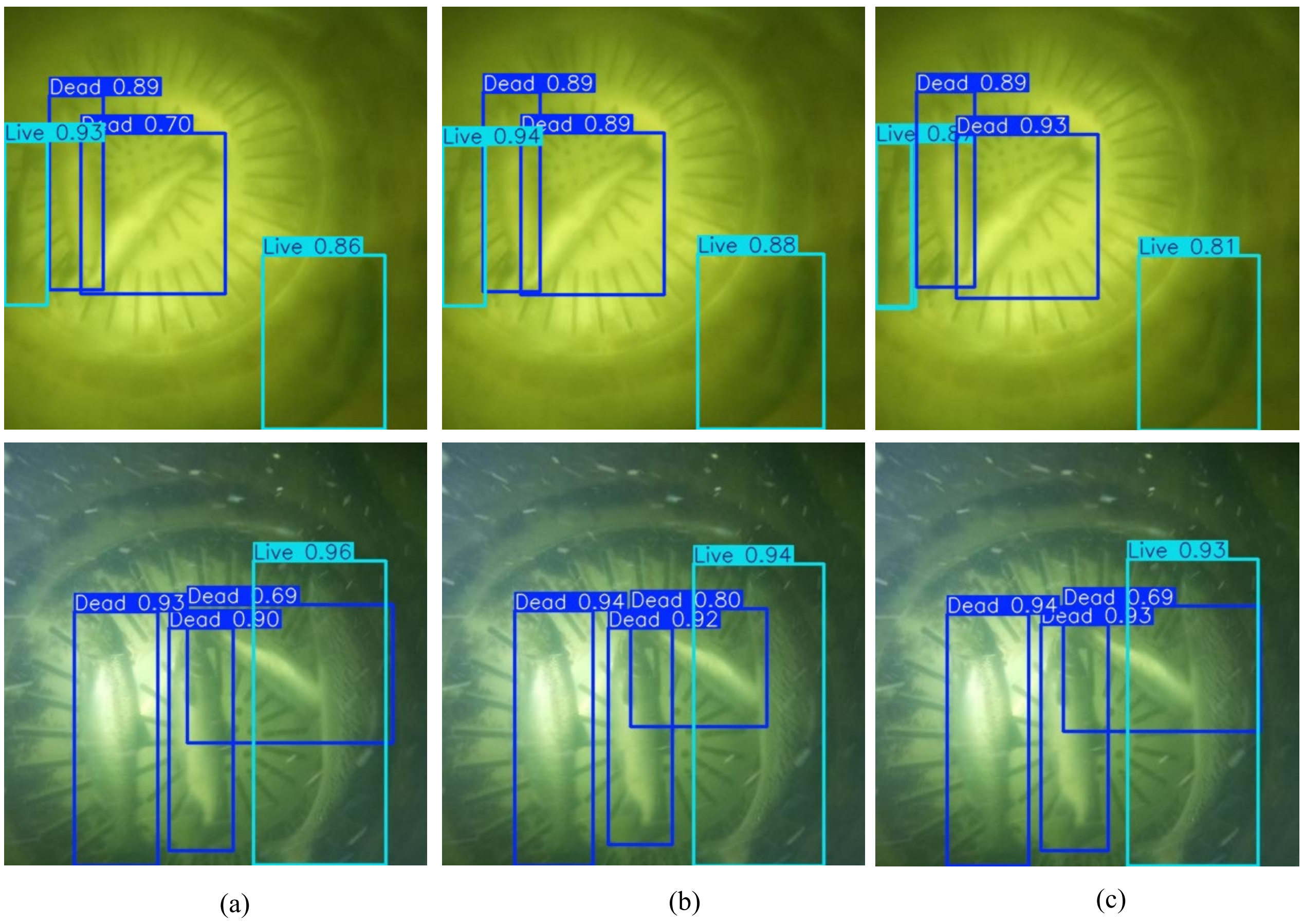

Figure 3. Representative dead and live fish detection output of (a) YOLO26n, (b) YOLO26s, and (c) YOLO26m, along with their detection confidence in ambient (top) and supplementary (bottom) light conditions. All three model variants successfully detected both dead and live fish under both lighting conditions, with confidence ranging for 0.69–0.94.

### 3.2 GPU inference speed

YOLO26 demonstrated the lowest postprocessing (*Post*, ms) latency across all three size tiers. In the nano tier, YOLO26n achieved the fastest post-processing (0.179 ms), accounting for only 18.0% of the combined inference and post-processing time. In comparison, YOLOv5nu reported 0.706 ms (44.2%), YOLOv8n recorded a markedly higher post-processing time of 1.805 ms (67.8%), and YOLO11n reported 0.702 ms (47.1%), indicating substantially higher post-processing overhead in the latter models relative to YOLO26n. The postprocessing overhead of YOLO26n was 10.1 times lower than that of YOLOv8n. A similar advantage was observed in the medium tier, where YOLO26m achieved a near-negligible post-processing time (0.128 ms), accounting for only 4.2% of the combined inference and post-processing budget. Compared to YOLO11m, which required 1.587 ms (40.2%), and YOLOv8m, which required 0.840 ms (27.5%), YOLO26m exhibited a substantially lower overhead. YOLO26m exhibited a 6.6 times lower post-processing overhead than YOLOv8m. Similarly, for the small tier, the post-processing overhead of YOLO26s was 1.7 times lower than YOLOv8s. Such efficiency gain can be attributed to YOLO26's NMS-free end-to-end inference design, which handles duplicate suppression within the network, eliminating the need for an explicit non-maximum suppression step and thereby reducing post-processing complexity and latency. Unlike post-processing time, the raw forward-pass inference time (*Infer*, ms) shows no consistent trend across model architectures and tested tiers. YOLO11n achieved the fastest forward-pass inference of 0.787 ms, followed closely by YOLO26n (0.813 ms) in the nano tier. In the medium tier, YOLOv5mu was fastest at 2.026 ms, while YOLO26m was slowest (2.919 ms), attaining a 44% longer forward pass relative to YOLOv5mu.

Despite YOLO26's NMS-free design, resulting in low post-processing overhead, this architecture had the highest end-to-end average latency (ms) and the lowest inference speed (FPS) across all size tiers. In the nano tier, YOLO26n had an average latency of 11.20 ms, resulting in an end-to-end inference speed of 89.31 FPS, compared to YOLOv8n at 9.18 ms (108.93 FPS), YOLOv5nu at 9.67 ms (103.41 FPS), and YOLO11n at 10.46 ms (95.60 FPS). The small and medium tiers followed the same trend. YOLOv8 consistently achieved the highest FPS and lowest average latency across all the size tiers. This discrepancy can be explained by examining the relationship between the measured inference components [i.e., forward-pass inference time (*Infer*) + Postprocessing time (*Post*)] and the reported average end-to-end latency (*Avg*). Beyond the *Infer* and *Post*, the *Avg* includes residual latency components related to additional pipeline overhead, such as image preprocessing, data loading, tensor allocation, and GPU-to-CPU memory transfers. YOLO26 variants incur the highest residual latency (i.e., *Avg - Infer - Post*) across all tiers, with the values of 10.21 ms (nano), 9.27 ms (small), and 8.52 ms (medium), compared to YOLOv8 at 6.52 ms, 6.83 ms, and 7.03 ms, respectively. This elevated residual overhead directly contributes to the higher end-to-end latency observed in YOLO26, offsetting the gains achieved through reduced post-processing. While the elimination of NMS reduces post-processing cost, YOLO26's one-to-one assignment head introduces additional tensor allocation and output decoding operations within the inference pipeline, which contribute to higher end-to-end average latency relative to NMS-dependent architectures on GPU hardware [35,37].

Notably, all twelve models comfortably exceeded 30 FPS, the widely referenced minimum threshold for real-time video processing [21], by a margin of 2.8 to 3.8 times.  For RAS mortality monitoring, where a fixed overhead camera observes a single tank and mortality events develop over minutes to hours, a detection rate of 5–10 FPS should be operationally sufficient for timely alert generation. All tested architectures, regardless of tier, provide substantially higher GPU throughput. While such high inference speeds achieved on high-end GPUs may be advantageous for more temporally demanding aquaculture tasks (e.g., feed waste monitoring, where small, fast-moving pellets require fine-grained temporal resolution and faster response for feeding optimization), the primary design constraint for aquaculture mortality monitoring shifts from maximizing FPS to optimizing cost efficiency, energy consumption, and deployment scalability. Therefore, low-power edge accelerators or CPU-based embedded systems may represent more practical and resource-efficient deployment choices without compromising operational effectiveness.

Table 4. GPU inference speed and computational efficiency for YOLO models trained on the full dataset (2,800 images)

| **Model** | **Tier** | ***Infer* (ms)** | ***Post* (ms)** | ***Avg* (ms)** | **GPU FPS** | **PT (MB)** | **Params (M)** | **Train (hrs)** |
|---|---|---|---|---|---|---|---|---|
| YOLO26n | Nano | 0.813 | **0.179** | 11.2 | 89.31 | **5.4** | 2.5 | 0.396 |
| YOLO11n | | **0.787** | 0.702 | 10.46 | 95.6 | 5.5 | 2.6 | 0.328 |
| YOLOv8n | | 0.857 | 1.805 | **9.18** | **108.93** | 6.3 | 3 | **0.312** |
| YOLOv5nu | | 0.892 | 0.706 | 9.67 | 103.41 | 5.3 | 2.5 | 0.318 |
| YOLO26s | Small | 1.472 | **0.461** | 11.2 | 89.15 | 20.3 | 9.9 | 0.478 |
| YOLO11s | | 1.78 | 0.792 | 10.91 | 91.66 | 19.2 | 9.4 | 0.373 |
| YOLOv8s | | **1.23** | 0.791 | **8.85** | **112.99** | 22.5 | 11.1 | **0.348** |
| YOLOv5su | | 1.539 | 0.523 | 9.46 | 105.71 | **18.5** | 9.1 | 0.353 |
| YOLO26m | Medium | 2.919 | **0.128** | 11.57 | 86.45 | 44 | 21.8 | 0.693 |
| YOLO11m | | 2.365 | 1.587 | 11.8 | 84.77 | **40.5** | 20.1 | 0.592 |
| YOLOv8m | | 2.214 | 0.84 | **10.08** | **99.16** | 52 | 25.9 | 0.524 |
| YOLOv5mu | | **2.026** | 1.229 | 10.72 | 93.3 | 50.5 | 25.1 | **0.514** |

*Infer (ms): raw GPU forward pass inference time | Post (ms): postprocessing time | Avg (ms): end-to-end average latency (ms) on A100 GPU | GPU FPS = end-to-end inference time (frame per second) on A100 GPU | PT (MB): PyTorch model size in Megabytes | Params (M) : Number of model parameters in millions*

### 3.3 Computational efficiency

Model efficiency (unitless), evaluated as mAP50 per megabyte of model size to investigate accuracy-to-storage trade-off, revealed substantial variation across architectures and tiers. The nano tier achieved the highest efficiency (15.0–17.8), followed by the small (4.2–5.1) and medium (1.8–2.3) size tiers (Tables 3 and 4). Therefore, nano-tier models may be best suited for edge devices with storage or memory constraints, such as microcontrollers or IoT devices with limited flash storage. For the nano tier, Yolov5nu (PT model size = 5.3 MB; mAP50 = 94.1%; number of parameters = 2.5M) achieved the highest efficiency. However, maximizing model efficiency does not necessarily align with all deployment objectives. In scenarios where storage is not a constraint and peak detection accuracy is prioritized, larger models such as YOLO11m (PT size = 40.5 MB; mAP50 = 94.81%; 20.1 M parameters) may be preferable due to their slightly higher mAP. A similar trend was observed for training efficiency, with nano-tier models requiring the least training time, followed by the small and medium tiers. Across all tiers, YOLO26 variants did not demonstrate advantages in either model efficiency or training efficiency relative to earlier YOLO architectures.

### 3.4 Learning curve analysis

Learning curve analysis revealed a consistent logarithmic improvement in mAP50, with the majority of performance gains occurring between 100 and 1000 training images across the developed models (Figure 4). Of the 12 models evaluated, four achieved mAP50 ≥ 90% with only 400 training images, while six reached this threshold with 700 training images (Figure 5). The YOLOv8 architecture demonstrated the highest training efficiency, as all variants (YOLOv8n, YOLOv8s, and YOLOv8m) attained ≥90% mAP50 with as few as 400 training images. In contrast, none of the YOLO26 models reached this performance level at 400 images. The YOLOv5u architecture showed comparable efficiency, requiring 400 to 700 images depending on the model size. Beyond 1,000 images, the model performance gains across all tested models were marginal. This trend is consistent with COCO-pretrained transfer learning models applied to domain-specific datasets, wherein pretrained feature representations capture a substantial portion of the target domain's visual structure with limited fine-tuning data. Therefore, additional training images primarily refine decision boundaries rather than learning new feature representations [10,38]. Overall, YOLO26 was the least data-efficient, as their nano (i.e., YOLO26n) and small (i.e., YOLO26s) tier models required 1,000 training images to surpass the ≥90% mAP50 threshold, approximately 2.5 times bigger dataset compared to their YOLOv8 counterparts. Only the medium tier of YOLO26 could reach this performance benchmark with 700 images.

While performance improvements beyond 1,000 training images were marginal, the associated training time increased substantially. For example, the YOLOv8 models achieved only a 1.8–2.5% increase in mAP50 when trained on the full dataset (2,800 images) relative to 1,000 images; however, training time increased by approximately 2.2–2.4 times across model size tiers. A similar trend was observed for other models. Notably, the time required to train the YOLO26 architecture was consistently higher than that of its predecessor models across all size tiers [Figure 4(b, d, f)]. The superior data and training efficiency of YOLOv8 can be attributed to its anchor-free, decoupled detection head, which simplifies label assignment and improves optimization of classification and localization tasks. These architectural advancements contribute to faster convergence and improved performance, particularly under limited data conditions [34,37]. In contrast, the one-to-one label assignment strategy adopted in YOLO26 reduces redundant predictions but yields sparser supervisory signals during training, slowing convergence and increasing the volume of training data required to achieve comparable detection performance [35]. Furthermore, the computationally intensive one-to-one matching during label assignment contributes to longer training times relative to architectures employing many-to-one assignment strategies, as reflected in the higher training durations observed for YOLO26 variants across all size tiers in this study.

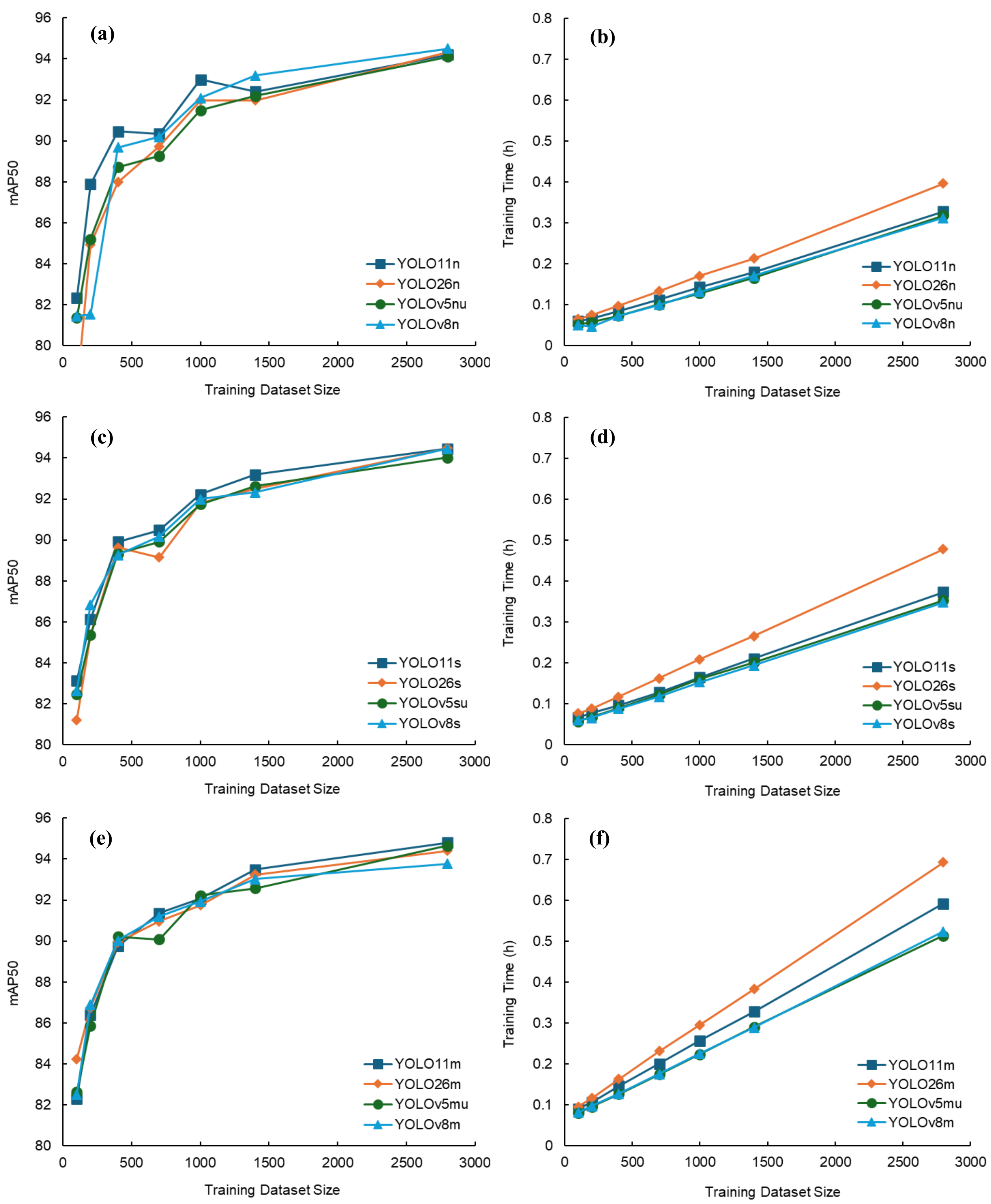

Figure 4. The learning curve analysis output illustrating the (a,c,e) mAP50 performance and (b,d,f) training time (hours) of YOLOv5u, YOLOv8, YOLO11, and YOLO26 across nano-, small-, and medium-tier models as a function of increasing training dataset size.

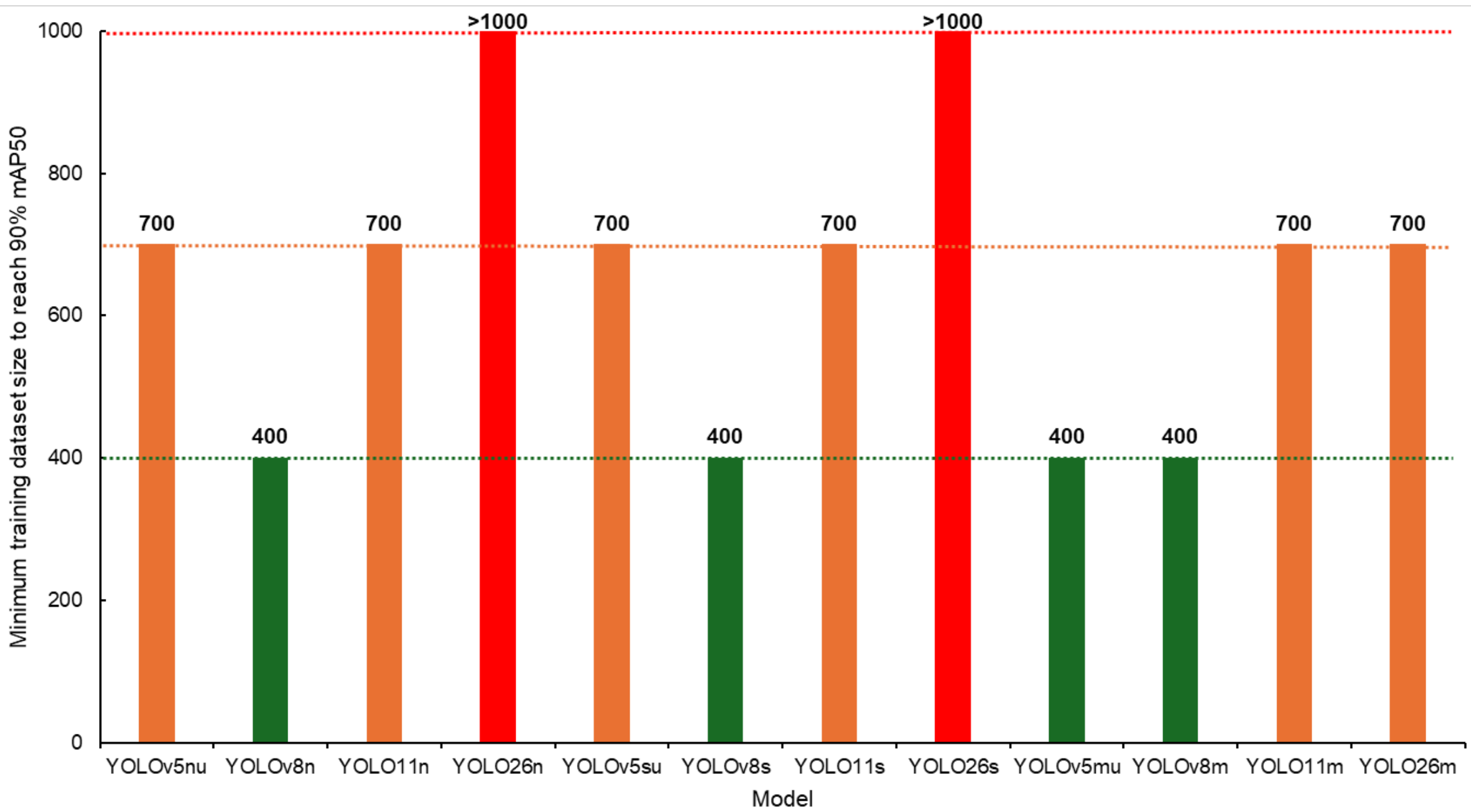

Figure 5. Minimum number of training images required to achieve ≥90% mAP50 on the test set across different model variants. Models highlighted in green achieved this performance threshold with as few as 400 training images; those in orange required approximately 700 images, while models shown in red did not reach ≥90% mAP50 even with up to 1,000 training images.

Overall, the observed logarithmic saturation, coupled with the increased training time associated with larger datasets, indicates that, despite the availability of extensive training data at the farm level, further annotation yields diminishing returns in accuracy. Therefore, additional resources may be more effectively allocated to improving image quality, increasing the diversity of mortality scenarios by collecting data for entire growth stages, or enhancing deployment infrastructure. Additionally, while the latest YOLO models promise deployment advantages for edge-devices, these gains come at the cost of reduced training efficacy and higher data requirements, suggesting that architectural generation should not be the sole criterion for model selection in aquaculture applications.

### 3.5. Edge deployment performance

Initial deployment trials on the Raspberry Pi 5 without active cooling resulted in CPU throttling within minutes of sustained inference, with peak temperatures above 80 °C, indicating thermal soft-limit conditions that reduce the CPU operating frequency below the nominal 2.4 GHz. Following the active cooling integration described in section 2.5, CPU temperature remained stable throughout all benchmark sessions (idle: 44.4 °C; peak: <70 °C), and no throttling was observed (throttle status: 0x0), confirming unthrottled operation at 2.4 GHz across all reported benchmark runs.

All nano-tier models achieved inference speeds exceeding 6 FPS on the Raspberry Pi 5 (Table 5), which provides sufficient temporal resolution for near-real-time mortality detection in RAS, where mortality events develop over minutes rather than seconds. YOLO26n achieved the highest inference speed in the nano tier at 133.2 ms per image (7.51 FPS), outperforming YOLOv5nu (7.19 FPS), YOLO11n (6.71 FPS), and YOLOv8n (6.51 FPS) by 4.5%, 11.9%, and 15.4%, respectively. This finding empirically validates YOLO26's claimed CPU inference advantages [32] in a domain-specific aquaculture deployment context, extending this claim beyond general-purpose COCO benchmarks. The observed speed advantage of YOLO26 can be primarily attributed to the elimination of NMS and DFL postprocessing steps, which reduce CPU computational burden and yield faster, more consistent per-image inference compared to NMS-dependent architectures [34,35,39]. Although inference was performed using ONNX Runtime, the observed 11.9% improvement over YOLO11n was considerably smaller than the up to 43% CPU ONNX inference speedup reported by Ultralytics. This difference is likely attributable to the use of a resource-constrained Raspberry Pi 5 ARM processor rather than the Intel Xeon platform used in the official benchmark, as well as differences in deployment environment, model optimization, and application workload.

Similar trends were observed for the small tier. YOLO26s achieved the top inference speed of 345.2 ms per image (2.90 FPS), followed by YOLOv5su (2.87 FPS), YOLO11s (2.79 FPS), and YOLOv8s (2.50 FPS). The most unexpected finding from the edge inference benchmark emerged in the medium tier. Unlike the consistent dominance of YOLO26 in the nano and small tiers, YOLOv5mu substantially outperformed all medium-scale architectures, achieving 1.25 FPS (797.6 ms) compared to YOLOv8m (1.06 FPS), YOLO26m (1.05 FPS), and YOLO11m (1.04 FPS). The superior ONNX runtime performance of YOLOv5mu on ARM CPU hardware can be attributable to the structural simplicity of YOLOv5's anchor-based convolutional architecture, which generates a computationally efficient ONNX inference graph that is well-suited to ARM processor instruction sets, with fewer operator types and no complex attention or one-to-one assignment mechanisms that introduce additional computational overhead on CPU hardware [34,40]. The inference speeds of small and medium-tier models (1.04–2.90 FPS) indicate that these architectures are better suited for periodic snapshot-based inspection at intervals of several seconds to minutes rather than continuous near-real-time monitoring on resource-constrained CPU-based edge hardware. Despite this, the edge-inference speed of all the tested models remains operationally viable for the mortality monitoring task. For tasks that require higher inference speeds, such as feed waste monitoring or behavioral analysis, GPU-accelerated embedded platforms may be a more suitable deployment solution.

ONNX model size across all tiers correlated closely with parameter count, with YOLOv8 consistently producing both the highest parameter count and the largest ONNX exports in every tier. YOLO26n produced the smallest ONNX export at 9.8 MB across all twelve models. For the Raspberry Pi 5 with 8 GB RAM used in this study, all twelve models loaded within available memory without issue. However, for deployment on more constrained embedded hardware with limited storage or RAM, the compact ONNX footprints of YOLO26n (9.8 MB) and YOLO11m (80.4 MB) at their respective tiers offer practical advantages in memory allocation overhead and model initialization speed for storage-limited on-farm edge devices.

Table 5. Raspberry Pi 5 inference benchmark results for all 12 YOLO models using ONNX format on 200 test images

| **Model** | **Tier** | ***Avg* (ms)** | ***Min* (ms)** | ***Max* (ms)** | ***P90* (ms)** | **CPU FPS** | **ONNX (MB)** | **Relative FPS % (vs. YOLOv8)** |
|---|---|---|---|---|---|---|---|---|
| YOLO26n | Nano | 133.2 | 124.9 | 260.3 | 145.1 | **7.51** | **9.8** | +15.4% |
| YOLO11n | | 149 | 143.7 | 170.1 | 152.3 | 6.71 | 10.6 | +3.1% |

| YOLOv8n | | 153.6 | 149 | 200 | 156.1 | 6.51 | 12.3 | Baseline |
|---|---|---|---|---|---|---|---|---|
| YOLOv5nu | | 139.2 | 133.9 | 183.8 | 140.9 | 7.19 | 10.3 | +10.4% |
| YOLO26s | Small | 345.2 | 340.9 | 393.6 | 349.8 | **2.9** | 38.2 | +16.0% |
| YOLO11s | | 358.9 | 352.5 | 408.7 | 364.8 | 2.79 | 37.9 | +11.6% |
| YOLOv8s | | 399.5 | 393.5 | 476.4 | 407.2 | 2.5 | 44.7 | Baseline |
| YOLOv5su | | 348 | 340.3 | 415.4 | 358.7 | 2.87 | **36.7** | +14.8% |
| YOLO26m | Medium | 956.4 | 948.4 | 998.7 | 969.2 | 1.05 | 81.7 | -0.9% |
| YOLO11m | | 959.5 | 950.2 | 1009.8 | 972 | 1.04 | **80.4** | -1.9% |
| YOLOv8m | | 944.7 | 935.8 | 988.8 | 956.7 | 1.06 | 103.6 | Baseline |
| YOLOv5mu | | 797.6 | 788.3 | 934.7 | 807.4 | **1.25** | 100.5 | +17.9% |

*Avg (ms): Mean end-to-end inference latency per image averaged over 200 test images | Min (ms): Minimum single-image inference latency observed across 200 test images | Max (ms): Maximum single-image inference latency observed across 200 test images | P90 (ms): 90th percentile inference latency, representing the worst-case latency experienced by 90% of processed images | CPU FPS: Frames per second achieved on Raspberry Pi, computed as 1000 / Avg (ms) | ONNX (MB): File size of the exported ONNX model used for Raspberry Pi 5 inference | Relative FPS % : Percentage improvement in FPS relative to the same-tier YOLOv8 baseline*

Overall, the latest YOLO26 architecture met several of its intended design objectives, though the benefits were selective for this domain-specific benchmark study. The NMS-free and DFL-free design of YOLO26 produced substantial improvements in CPU inference performance on the Raspberry Pi 5. Among the nano and small model tiers, YOLO26n achieved the highest inference speed, validating its suitability for faster CPU-based deployment in practical aquaculture applications. In addition, the compact ONNX export size of YOLO26 supports deployment under resource-constrained on-farm conditions. However, the same NMS-free one-to-one assignment mechanism that improved CPU inference efficiency also introduced training-related trade-offs. The YOLO26 nano and small variants required approximately 2.5 times more annotated images than YOLOv8 to surpass the 90% mAP50 threshold and consistently incurred higher training-time costs. In contrast, YOLOv8 demonstrated balanced performance across the entire deployment pipeline. The model converged with comparatively limited training data, delivered strong GPU throughput, and transferred efficiently to the ONNX runtime without the pipeline overhead penalties observed in YOLO26. YOLOv8 also exhibited consistent behavior across both GPU and CPU deployment environments. While the anchor-free, decoupled head of YOLOv8 resulted in larger ONNX export sizes, this model produced a stable, well-optimized inference graph with predictable runtime characteristics. These advantages have likely contributed to the broader adoption of YOLOv8 in aquaculture applications [29], and in other computer vision domains. The architectural trade-offs among YOLO generations are substantial, and model selection should be guided by application-specific constraints, including annotated data availability, deployment hardware, and real-time inference requirements.

## 4. Conclusion

The research question motivating this benchmarking study, whether the NMS-free architecture of YOLO26 represents a meaningful advancement over earlier YOLO generations for fish mortality detection in aquaculture, does not yield a definitive answer. The findings demonstrate that architectural generation alone is an insufficient criterion for model selection, and architectural novelty does not directly translate to model superiority in domain-specific aquaculture applications. All 12 trained models across the architectural and size tiers converged within 1.04 percentage points of mAP50 for mortality detection. Therefore, the primary differentiators between the models should not be detection accuracy, but deployment-related factors such as the availability of annotated data, computational hardware constraints, and inference speed requirements. Additionally, findings from this benchmark study highlight an important limitation of relying solely on GPU-based benchmarking as a proxy for edge deployment performance. The inference speed rankings observed on the A100 GPU did not translate to the CPU-based Raspberry Pi 5 and would have led to suboptimal model selection if GPU performance had been used as the sole deployment criterion. These results reinforce the need for deployment-oriented evaluation on target edge hardware in future studies, as processor architecture, memory constraints, and runtime-specific graph optimization can substantially affect edge performance relative to GPU benchmarks.

Although architectural advancements contribute incremental improvements in detection performance, greater emphasis should be placed on the more fundamental challenges of developing diverse, real-world training datasets and conducting rigorous optimization and validation under commercial-scale aquaculture conditions. For this benchmark study, sufficient training images were collected during the grow-out stage under two lighting conditions over a 90-day period; however, the dataset lacked mortality data from juvenile production phases and represented only a single grow-out tank configuration. Additionally, the generalizability of the models to other species, facility designs, or production environments remains unverified. In addition, the edge inference benchmark was run on a single Raspberry Pi 5 device and evaluated exclusively with ONNX Runtime. Alternative deployment frameworks, including TensorFlow Lite (TFLite) and OpenVINO, were not assessed on multiple edge devices in this study. Therefore, the findings of this study should be interpreted as context-specific, evidence-based guidance rather than universally generalizable deployment recommendations.

Future work should expand benchmarking efforts across various aquaculture production systems (e.g., net-pen, RAS), species, and growth stages to establish the cross-context generalizability required for industry-wide deployment recommendations. Furthermore, evaluation of the developed models on GPU-accelerated embedded platforms, such as the NVIDIA Jetson series, would provide a more comprehensive understanding of performance across diverse edge hardware environments and may reveal inference advantages of newer architectures that were not fully apparent under the CPU-constrained conditions evaluated in this study.

## CRediT authorship contribution statement

Rakesh Ranjan: Conceptualization, Methodology, Investigation, Formal analysis, Data curation, Funding acquisition, Writing – original draft; Gajanan Kothawade: Writing – review & editing, Data curation; Kata Sharrer: Writing – review & editing, Data curation; Scott Tsukuda: Writing – review & editing, Data curation; Christopher Good: Writing – review & editing, Funding acquisition.

## Funding

This study was supported by the USDA Agricultural Research Service agreement numbers 59-8082-5-002. Use of tradenames is solely to provide accurate information and does not imply endorsement by the USDA. The Conservation Fund and USDA are equal opportunity providers and employers. All experimental protocols and methods were approved by the Freshwater Institute's Institutional Animal Care and Use Committee.

## Acknowledgments

The author extends their sincere thanks to Brian Vinci and Travis May for their valuable suggestions and assistance with mortality data collection.

**Declaration of competing interest**

The authors declare that they have no known competing financial interests or personal relationships that could have appeared to influence the work reported in this paper.

**Use of AI tools declaration**

During the preparation of this work, Generative AI tools (Claude AI and ChatGPT) were used solely to improve the readability and language of the manuscript. All AI-assisted revisions were thoroughly reviewed and edited for accuracy and relevance.

**Data availability**

The fish mortality dataset (RAS-MortDB) used in this study, along with the trained model weights in PyTorch (.pt) and ONNX (.onnx) formats for all twelve evaluated architectures, will be publicly available on the GitHub repository.